\documentclass[runningheads]{llncs}
\usepackage[T1]{fontenc}
\usepackage{graphicx}
\usepackage{xcolor}
\usepackage{array}
\usepackage{hyperref}
\usepackage{subcaption}

\DeclareRobustCommand{\sombrero}{\raisebox{-0.2\height}{\includegraphics[height=1.1em]{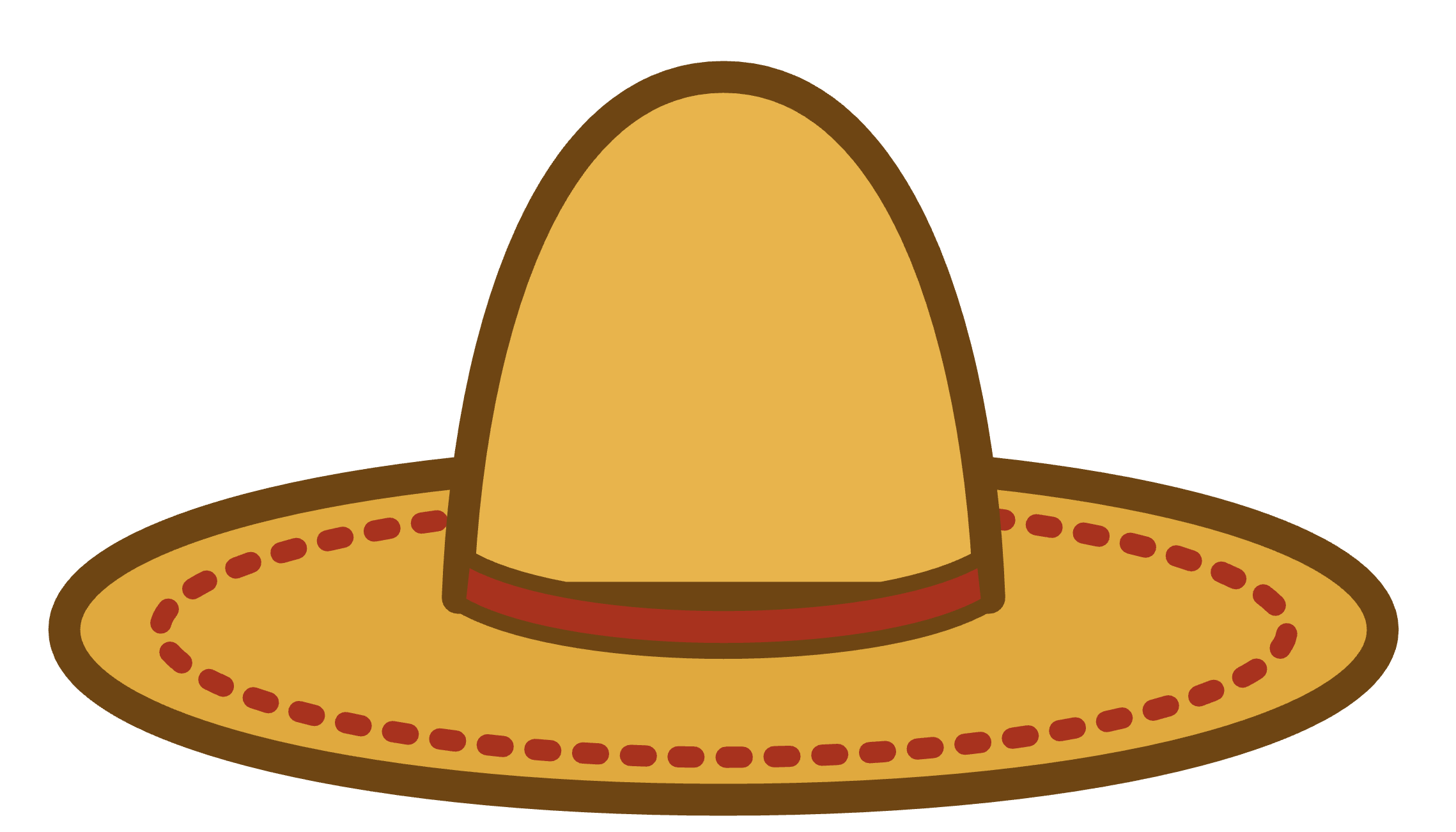}}}

\begin{document}
\title{\sombrero~MexHat: A Dataset for Hate Speech Detection in Mexican Spanish Videos\thanks{Preprint submitted to CIARP 2026}
}
\titlerunning{MexHat: A Dataset for Hate Speech Detection in Mexican Spanish Videos}
\author{Itzel Tlelo-Coyotecatl\inst{1} \and Hugo Jair Escalante\inst{1,2}}
\authorrunning{Tlelo-Coyotecatl and Escalante}
\institute{$^1$INAOE, Puebla, Mexico\\  $^2$The University of Texas at El Paso, El Paso, TX, USA\\ \email{\{itlelo,hugojair\}@inaoep.mx}
}
\maketitle % typeset the header of the contribution
\begin{abstract}
Ensuring online safety through content monitoring had raised Hate Speech Detection as a crucial task to be addressed.
By essence the task demands the capture of contextual cues, which are essential for a precise understanding of the content's intent.
Although automated detection approaches for the task have advanced significantly, the scarcity of non-English resources persists, limiting the ability of models to adapt to the subtle, context-dependent, and culturally related nature of multimodal content.
In this paper, we introduce MexHat, a video dataset designed to capture the linguistic and cultural cues for the hate-speech detection task in a Mexican Spanish context.
Our dataset comprises around 1k video clips annotated across two tasks: a three-way class evaluation (no negative content, offensive content and hate-speech content), and a fine-grained class evaluation including three hate-speech sub-categories.
The dataset statistics and the baseline results highlight the inherent challenges associated with the task. 
Disclaimer: This paper contains sensitive content that may be disturbing to some readers.
\end{abstract}
%
%
% --------------------------------------------
\section{Introduction}
Nowadays, the generation and consumption of a variety of media content is inevitable. 
Unfortunately, part of that content could be produced with malicious intent.
Given the ease of content propagation through social platforms, the development of automated approaches for detecting and monitoring harmful content has become a critical need in order to promote a safety environments between online communities.

Driven by the need to capture human behavior, multimodal processing has gained significant popularity across various automated detection tasks \cite{AffectComp:RecentAdvChallandFutTrends}. Particularly, tasks that involve language interpretation rely on the capture of multiple cues, such as linguistic expressions, tones, and body language. That is because language itself is deeply dependent on context. This requirement is evident in affective computing-related domains like sentiment analysis \cite{gandhi2023multimodalSA}, sarcasm detection \cite{Farabi2024ASO-SarcD-survey}, comic mischief detection \cite{baharlouei-etal-2024-labeling-comic}. Regarding this multi-cue approach, this work is aimed at the task of hate speech detection in video.

Although focused on video analysis, the hate speech detection approaches have recently adopted multimodal modeling. Early works, such as those by Shang et al. \cite{Shang2019VulnerCheck} and Wu \& Bhandary \cite{Wu&Bhandary2020DetectionHSinVusingML}, addressed the task using text-only features from transcripts or comments, a limited strategy when aiming to capture complex forms of hostile behaviors. By integrating audio and text cues, Rana \& Jha \cite{Rana2022EmotionBH} explored the capture of richer representations for the video content.
More recently, works like Das et al. \cite{Das2023HateMMAM} jointly fused textual, audio and visual information to capture complex dynamics between modalities. Consequently, as approaches integrate more modalities, the related challenges shifts toward how to adequately capture and model these interdependent hateful cues.

Beyond the computational stage challenges, capturing hateful cues derived from language requires not only contextual but cultural understanding. Because the majority of existing literature is biased toward English content, achieving data representativeness remains an open challenge that requires developing resources tailored to specific cultural targets.
Some efforts to cover this low-resource gap include the following works. Alcantara et al. \cite{Alcntara2020OffensiveVD} built up a Portuguese dataset by retrieving content from Youtube platform and addressing the task through a classic approach using text modality. Urbano et al. \cite{HernandezUrbanoJr2021ABH} also evaluated the text modality with a transfer learning approach processing around 1k of Tagalog-related videos. More recently, Wang et al. \cite{Wang2024MultiHateClipAM} chosed youtube and bilibili platforms to retrieve both English and Chinese content.
However, there is almost no presence for Spanish, causing a significant linguistic gap considering it as one of the top five\footnote{https://www.babbel.com/en/magazine/the-10-most-spoken-languages-in-the-world} most spoken languages in the world.

Motivated by the lack of specialized resources and the potential to enhance model adaptability across diverse contexts, our main contributions are:
\begin{itemize}
    \item We built MexHat, a Mexican-Spanish video dataset for hate-speech detection. This dataset includes a little more than 1k video clips and provides multimodal features (e.g., transcriptions, audio, video). To the best of our knowledge this is the only available resource for this language.
    \item We introduce a culturally-aware annotation framework that addresses a high-level three-way classification by targeting no negative content, offensive content, hate-speech content. Also, a fine-grained target classification between sub-categories of hate-speech was determined. Both tasks were proposed in order to explore context-dependent cues at different levels given the associated language and the culture.
    \item We introduce comprehensive baselines for both tasks that establish a new reference for content moderation and reflect the subjectivity an challenges of the task.
\end{itemize}

\section{Building the MexHat Dataset}
This section describes the video collection, filtering, and annotation processes employed to build our dataset, as well as, instance examples.

\subsection{Data Collection}
We choose the YouTube platform to filter and download videos for our dataset. Three ways of filtering were considered: 1) using hate-speech related words; 2) using relevant identified videos as seeds to retrieve similar content; and 3) identifying playlists of channels with possible hate-speech content.

First, two lists of hate-speech related terms were delimited and used to filter $n$ number of videos per term. One list was based on expressions and terms that should be avoided to use inclusive and non-sexist language. This list was based on the \textit{Guía de lenguaje incluyente y no sexista} \cite{SRE_GuiaLengIncluyenteynoSexista} published by the \textit{Secretaría de Relaciones Exteriores} including terms (56) that should be avoided like: \textit{amanerado, afeminado, machorra, sordito, sidoso, mojado}.
The other list was based on terms catalogued as offensive in the mexican context. The terms (29) were obtained from the \textit{hatebase\footnote{https://hatebase.org/} website}. Some examples include: \textit{amariconado, cholo, lagartona, maricon, joto, prieto}.

After retrieving the first pool of videos using the hate-speech related terms, we manually identified some relevant videos with possible hate speech content.
A video was considered relevant if it had a duration of maximum 20 minutes, and explicitly included offensive language or controversial topics. Then, mostly stand-up sketches, opinion videos and soap operas where considered.
By using the YouTube API, these video IDs served as seeds to retrieve similar content. 
Alongside this, we manually identified channels and playlists with possible hate speech content and retrieved some of the publicly available videos.
After downloading the videos, each video was segmented into 1-minute clips, resulting in a total of around $8$k video clips. 

\subsection{Annotation Process}
\label{subsection: annotation process}

\begin{figure}[h!]
\centering
% --- IMAGE A ---
\begin{subfigure}[b]{0.5\textwidth}
    \centering
    \includegraphics[width=\textwidth]{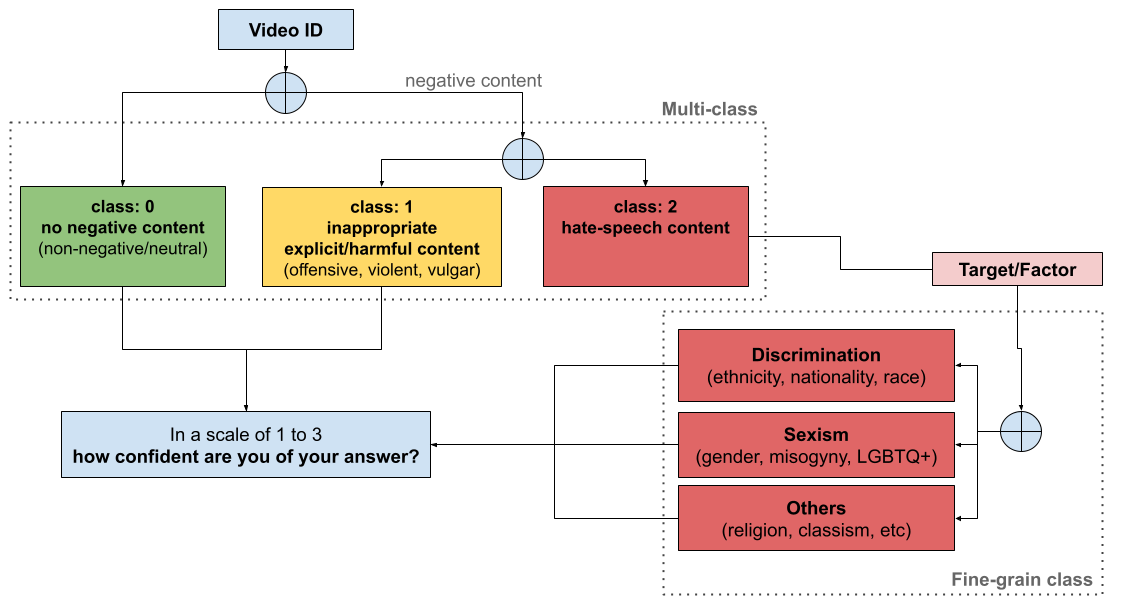}
    \caption{Simplified guideline scheme}
    \label{fig:dataset-guidelines}
\end{subfigure}
\hfill 
% --- IMAGE B ---
\begin{subfigure}[b]{0.48\textwidth}
    \centering
    \includegraphics[width=\textwidth]{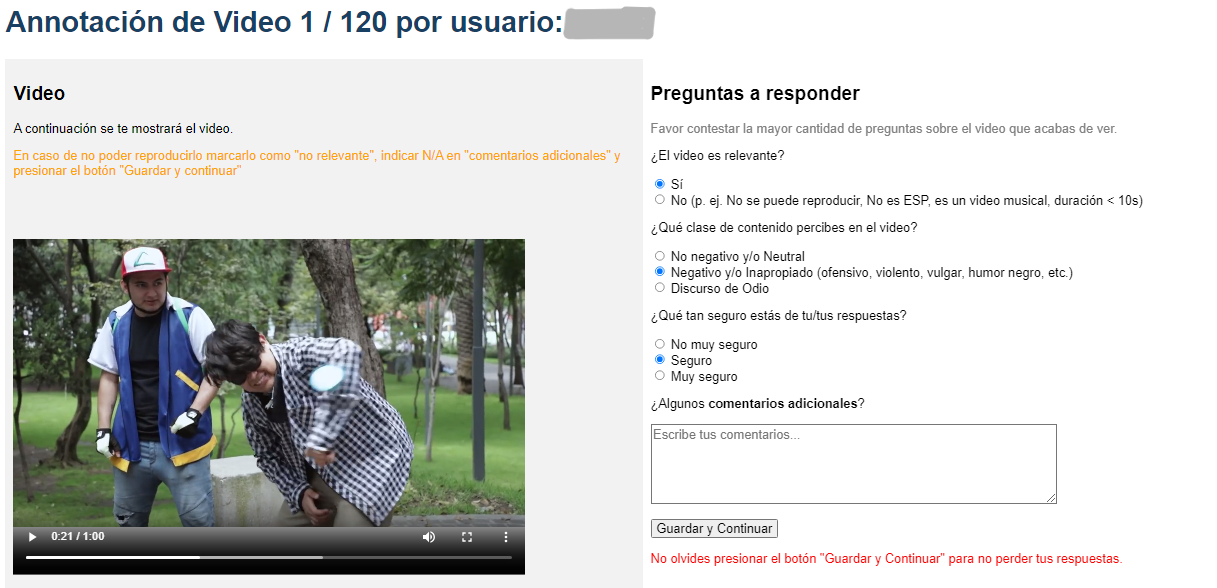} 
    \caption{Annotation platform}
    \label{fig:annotation-platform}
\end{subfigure}
\caption{Overview of the annotation process}
\label{fig: dataset-guidelines} 
\end{figure}

\begin{figure}[h!]
\centering
% --- IMAGE A ---
\begin{subfigure}[b]{0.43\textwidth}
    \centering
    \includegraphics[width=\textwidth]{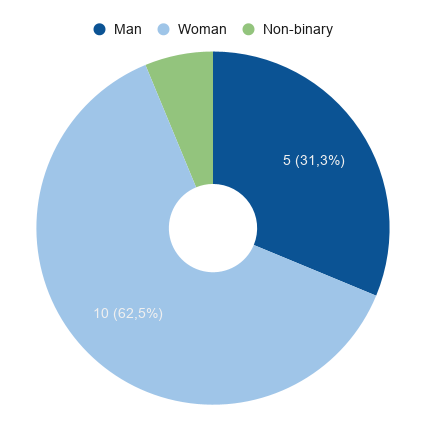}
    \caption{Gender Identity}
    \label{fig:profile-gender}
\end{subfigure}
\hfill 
% --- IMAGE B ---
\begin{subfigure}[b]{0.43\textwidth}
    \centering
    \includegraphics[width=\textwidth]{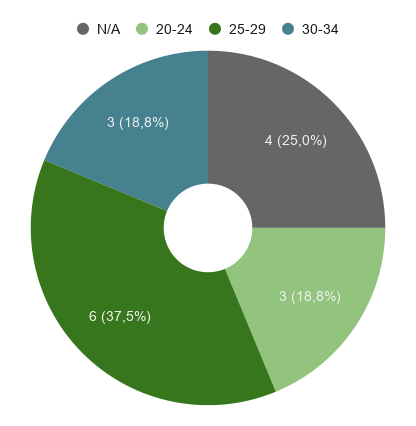} \caption{Age}
    \label{fig:profile-agerange}
\end{subfigure}
\\
% --- IMAGE C ---
\begin{subfigure}[b]{0.43\textwidth}
    \centering
    \includegraphics[width=\textwidth]{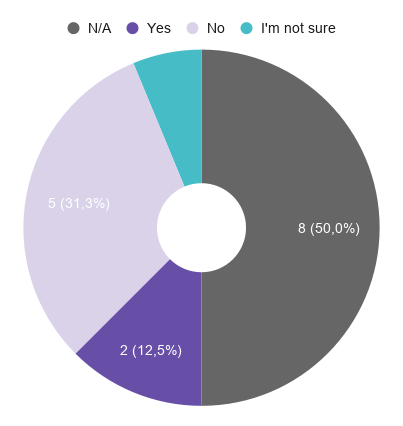}
    \caption{Do you identify as part of the LGBTQ+ community?}
    \label{fig:profile-community}
\end{subfigure}
\hfill 
% --- IMAGE D ---
\begin{subfigure}[b]{0.43\textwidth}
    \centering
    \includegraphics[width=\textwidth]{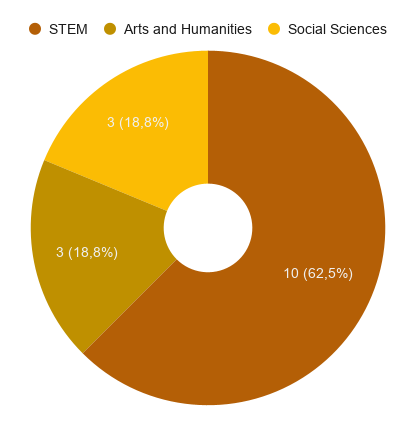} 
    \caption{Background}
    \label{fig:profile-background}
\end{subfigure}
\caption{Annotator demographics.}
\label{fig: dataset-profile-stats} 
\end{figure}

Annotators were provided with a guideline (See Figure \ref{fig: dataset-guidelines}) and asked to label the provided video clips into one of the three main categories: \textit{no negative content}, \textit{offensive/inappropriate content} and \textit{hate-speech content}. 
To identify \textit{hate-speech content} the definition provided by the United Nations\footnote{https://www.un.org/en/hate-speech/understanding-hate-speech/what-is-hate-speech} was considered. That is, \textit{“any kind of communication in speech, writing or behaviour, that attacks or uses pejorative or discriminatory language with reference to a person or a group on the basis of who they are, in other words, based on their religion, ethnicity, nationality, race, colour, descent, gender or other identity factor.”} Then, \textit{offensive/inappropriate content} involved harmful, offensive, violent and vulgar content that does not explicitly show a target group. Finally, all the videos that were not labeled in these clases were categorized as \textit{no negative content}. 
If a video clip was labeled as \textit{hate-speech content} we asked annotators to categorize it into one of three groups: (G1) comprising discrimination by ethnicity, nationality or race; (G2) including factors related to sexism, and attacks on the LGBTQ+ community, (G3) comprising all the hate forms that were not in the previous groups. Additionally, we required them to specify the \textit{level of confidence in their answers}. 

Sixteen annotators participated in the labeling process during three annotation campaigns. 
Each annotator was provided with batches of around $100$ up to $250$ video clips. Figure~\ref{fig: dataset-profile-stats} summarizes the demographics of annotators. 
Female participation and STEM background dominated, but social sciences, and arts and humanities were also present. 
Most participants were between 25 to 29 years old.
Also, while a few percentage of the group explicitly identified as part of the LGBTQ+ community, the majority did not provide that data.

\subsection{Annotator Agreement}
During the annotation phase (Section \ref{subsection: annotation process}) each video clip was provided to two independent annotators.
A total of $1540$ video clips were labeled. 
About $10.51\%$ were not included in the final dataset because both of the annotators or one of them considered the video clip as irrelevant. That is, the duration was less than 10 seconds, was not in Spanish or it only contained music and static background images.
Then, the dataset ended up with $1378$ videos.
For estimating annotators' agreement we calculated a $\kappa$ (\textit{linear Kappa}) statistic. 
For the three-class labeling we obtained $\kappa = 0.416$ 
and for the fine-grained labeling we got $\kappa = 0.514$. 
Both scores might be considered as moderated agreement given the subjectivity of the task.
To assign the final labels, in case of disagreement, the hierarchy of the chosen classes was considered. That is, hate-speech labels had priority over offensive labels and no negative content. This was in order guarantee available examples in each class.

\begin{figure}[h!]
\centering
\includegraphics[width=0.6\textwidth]{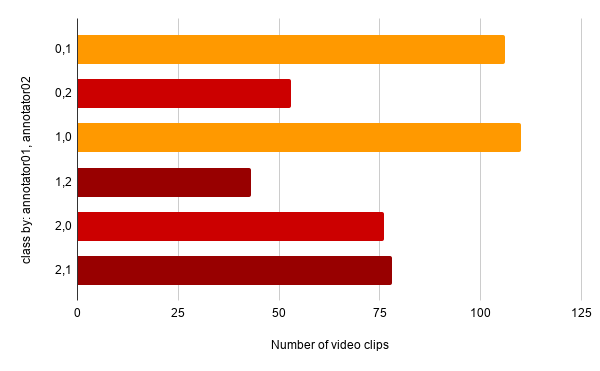}
\caption{Number of videos with disagreement at annotation stage.} \label{fig: dataset-annotators_disagreement}
\end{figure}

At the annotation stage, around 466 videos had different votes, that is, around $34\%$ percent of the final dataset. 
Figure \ref{fig: dataset-annotators_disagreement} shows that the most common disagreements were labeling a video into the no negative content (0) vs. the offensive content (1). 
Similar case disagreements between no negative content (0) vs. hate-speech content (2) occurred.
Considering  the variety of participant profiles, it is shown that there is still a thin line between what could be considered offensive or  not.
Additionally, disagreement between negative content, involving \textit{offensive content} vs. \textit{hate-speech content} were less evident but still represent a difficult task when trying to separate both concepts.

\subsection{Text Modality between Classes}
According to the obtained results (Section \ref{subsection: results}), some of the configurations with better performance involved the \textbf{T}ext modality.
To catch a glimpse of the difference in the vocabulary for the classes \textit{no negative content}, \textit{offensive content} and \textit{hate-speech content}, word clouds were generated after removing stopwords\footnote{Spanish stopwords provided in https://github.com/Alir3z4/stop-words.git}.

\begin{figure}[h!]
\centering
% --- IMAGE A ---
\begin{subfigure}[b]{0.32\textwidth}
    \centering
    \includegraphics[width=\textwidth]{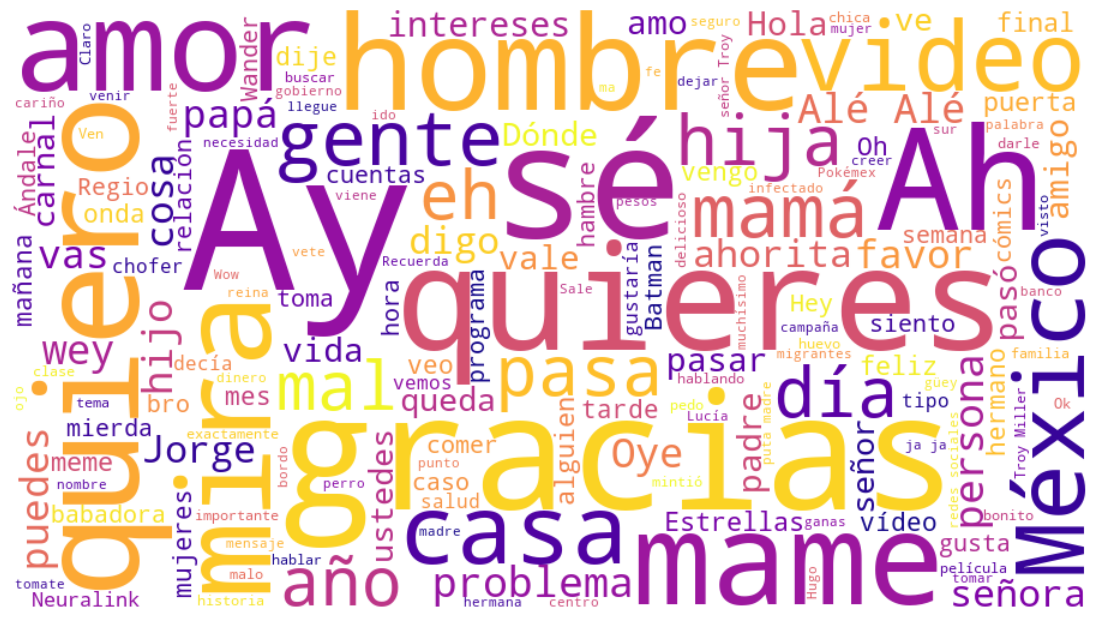}
    \caption{NN: no negative}
    \label{fig:wordcloud_NN}
\end{subfigure}
\hfill 
% --- IMAGE B ---
\begin{subfigure}[b]{0.32\textwidth}
    \centering
    \includegraphics[width=\textwidth]{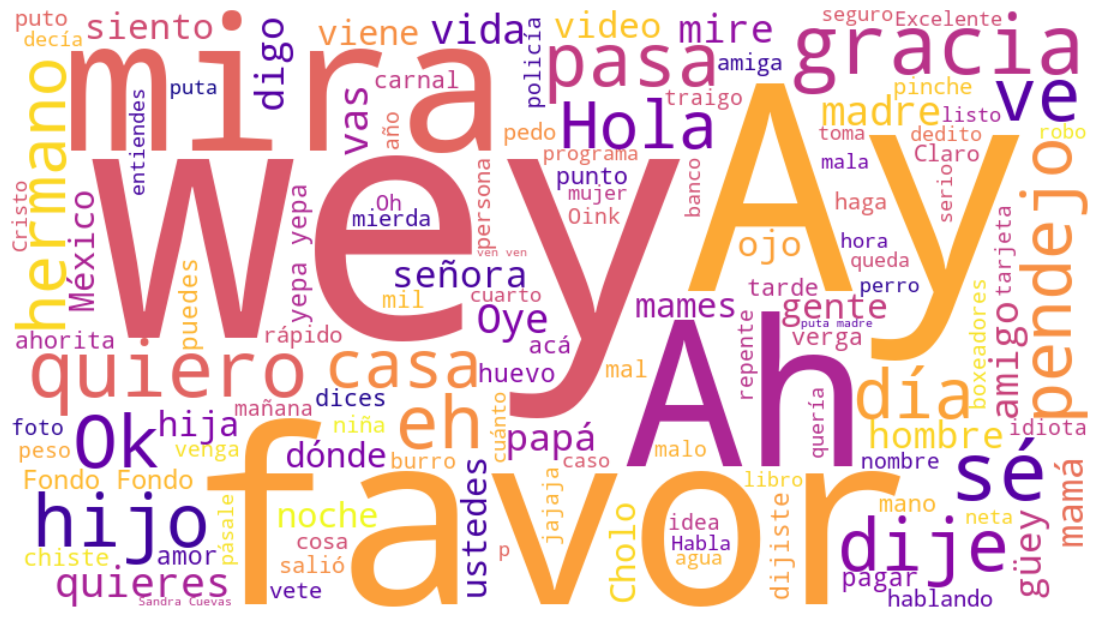}
    \caption{OF: offensive}
    \label{fig:wordcloud_OF}
\end{subfigure}
\hfill 
% --- IMAGE C ---
\begin{subfigure}[b]{0.32\textwidth}
    \centering
    \includegraphics[width=\textwidth]{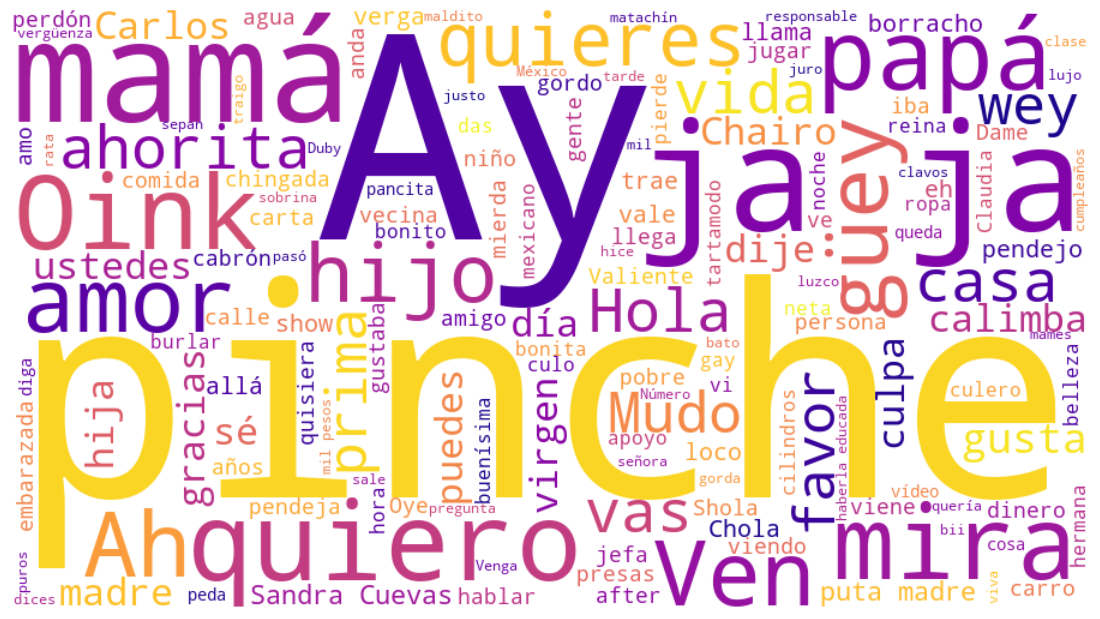} 
    \caption{HS: hate-speech}
    \label{fig:wordcloud_HS}
\end{subfigure}
\caption{Wordclouds from test partition transcriptions.}
\label{fig: wordclouds_threeclasses} 
\end{figure}

Figure \ref{fig: wordclouds_threeclasses} shows the unique representative words for each one of the main three evaluated classes. 
While class \textbf{NN} shows a greater variety of general terms, in \textbf{OF} class it is clear a high frequency of terms like: \textit{wey}, \textit{ay}, \textit{favor}, \textit{ah}, \textit{mira}. 
On the other hand, most frequent terms in \textbf{HS} class include: \textit{ja}, \textit{ay}, \textit{pinche}, \textit{mamá}, \textit{quiero}.
Considering these terms, \textit{wey} and \textit{pinche} could be considered as explicit ways to refer to someone and depending on the context could be interpreted as despective.

\subsection{Examples of Videos}
\begin{figure}[h!]
    \centering
    % ----------- SUBFIG 1
    \begin{subfigure}{\linewidth}
        \centering
        \includegraphics[width=0.82\linewidth]{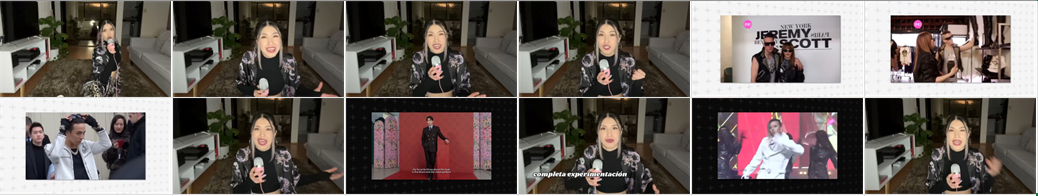}\\
        \tiny \textbf{Transcription:} Today, I'm feeling very second-generation. No, it's just horrible. The bolder and more modern you looked, the more it seemed like you were actually understanding the assignment. It is towards the end of the second generation where we see K-Pop's first approaches to luxury fashion. The first idols to catch the interest of these designers were none other than CL and G-Dragon. CL had a very close exchange with Jeremy Scott almost since 2NE1's debut, and she became one of his muses. In 2013, he designed the outfits for 2NE1's World Tour. G-Dragon, on his part, worked closely with Chanel until he became a brand ambassador. And from that point on, it’s history. Now, it is completely normalized, and we see idols everywhere acting as muses and ambassadors for various fashion brands. The second generation was a period of complete experimentation for K-Pop, resulting in the industry's boom. They wrote the rules and reconstructed the foundations for the future of the genre and the international exposure of Korean entertainment. In this video, I wanted to focus purely on touching upon some creative aspects, but on the commercial side, there is a whole world to explore.
        \caption{Video (MsAtrx97Uc8.15) from class \textbf{NN}.}
        \label{fig:nn_frames}
    \end{subfigure}
    \vspace{0.4cm} 
    % ----------- SUBFIG 2
    \begin{subfigure}{\linewidth}
        \centering
        \includegraphics[width=0.82\linewidth]{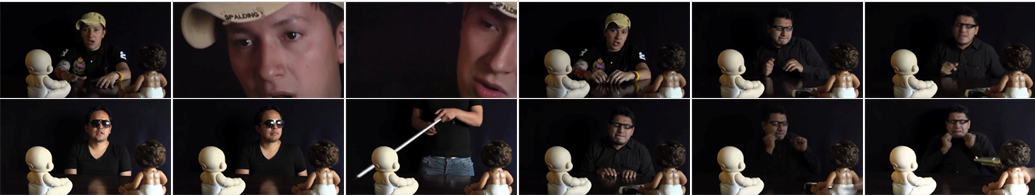}\\
        \tiny \textbf{Transcription:} Come on, dude, no way, calm the fuck down, that's why I don't give a shit about the baby stuff anymore, don't cry, well yeah it hits hard, but at the end of the day look, they are siblings, they can love each other like always, I don't have my boss with me, let's see kid, what color is shit, what color is crap? It's just that sometimes when I'm sick it comes out like that, son of a bitch, motherfucker... who wouldn't you trust? If you had to give someone a job, who would you give it to? Let's see kid, this one is the bad guy, this one has a gun, he's going to rob you, he's going to kill your family, all your friends, he's going to kill them with this gun, this is the color of God, the God you believe in, the one who's going to kill you, which one do you think? Which one do you think is going to rob you?\\
        \caption{Video (MbAcTRAYFS8.04) from class \textbf{OF}.}
        \label{fig:of_frames}
    \end{subfigure}
    \vspace{0.4cm} 
    % ----------- SUBFIG 3
    \begin{subfigure}{\linewidth}
        \centering
        \includegraphics[width=0.82\linewidth]{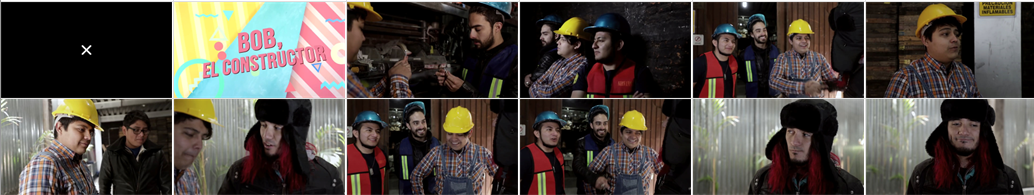}\\
        \tiny \textbf{Transcription:} Hi, I'm Bob, and together with my two assistants, Charlie and Bert, I dedicate myself to building, restoring, and repairing all kinds of things. It's hard work, but somebody's gotta do it. Bob the Builder. Today, a client came in looking to remodel their house. I already saw it, and I can tell you the result is going to look phenomenal. It's one of the easiest cases we've ever had, and thanks to my team, we'll be able to finish it in about two or three days. Today, I came with my friend Bob to have him check out my house because it needs some remodeling. Honestly, I trust his work a lot; he's the best. My house needs remodeling; it's in really bad shape, honestly, look. Wow! That is the shittiest house I have ever seen. Yeah, it really is a piece of shit. Not even pigs could live in there. How do you live in that pigsty? Even the color is hideous. Who painted it? A blind guy? Who was it, Shrek? Did a hurricane pass through there? Shrek lives in a swamp, and he lives better than you. Honestly, you seem dirt poor; I wonder if you even have enough to pay for the remodeling. But oh well, what do you say, boys? Can we fix it?\\ 
        \caption{Video (kK-3F5epeyE.00) from class \textbf{HS}.}
        \label{fig:hs_frames}
    \end{subfigure}
    %\vspace{0.3cm} 
    %
    \caption{Examples of videos from classes \textbf{NN}: no negative content, \textbf{OF}: offensive content, and \textbf{HS}: hate-speech content.}
    \label{fig: dataset-video-examples}
\end{figure}

Figure \ref{fig: dataset-video-examples} shows examples of videos~\footnote{ASR translations were generated with \textit{3.5 Flash} Gemini model.} labeled into the three main classes (NN, OF, HS). 
NN video (Figure \ref{fig:nn_frames}) shows a girl facing the camera and discussing a music-related topic.
For OF content (Figure \ref{fig:of_frames}), some men were interviewed regarding two dolls; during these sessions, the use of derogatory language was observed.
Finally, in HS content case (Figure \ref{fig:hs_frames}) the video shows a sketch in which a man asks for home renovation advice, and the expert characters mock him.
This video was labeled as a possible case of classism, hence it belongs to the third group of hate-speech (hs3: \textit{others}).

\subsection{Feature Extraction}
We extracted features for text, audio and visual modalities using pretrained models.
To obtain \textbf{text modality features}, audio was transcribed from the video using whisper\footnote{https://github.com/openai/whisper} (\textit{model=“medium”}). 
Then, transcribed text was processed with the BETO encoder~\cite{CaneteCFP2020BETO} ending up with representations of $n_{t} \times 768$ per video clip.
For the \textbf{video modality}, I3D representations for flow ($n_{v} \times 1024$) and rgb ($n_{v} \times 1024$) were obtained from each video clip~\cite{Carreira2017QuoVA-i3d}.
Finally, \textbf{Audio} features were obtained by using the vggish pretrained model~\cite{Hershey2016CNNAF-vggish} corresponding with $n_{a} \times 128$ representations per video clip.
Table~\ref{table: dataset-dim-statistics} summarizes the statistics on the extracted multimodal descriptors. 

\begin{table}[h!]
\caption{Statistics for each modality. Each video corresponds to a $n \times D$ representation. } \label{tab1}
\centering
\begin{tabular}{|l|c|r|r|r|r|r|r|}
\hline
Modality & $n_{min}$ & $n_{max}$ & $n_{avg}$ & $n_{median}$ & $n_{std}$ & $D$ \\
\hline
Audio& 2 & 69 & 59.27 & 62 & 10 & 128 \\
Text & 2 & 488 & 177.36 & 186 & 73.23 & 768\\
Video& 0 & 31 & 24.88 & 27 & 5.25 & 1024+1024\\
\hline
\hline
\end{tabular}
\label{table: dataset-dim-statistics}
\end{table}

Given the varying number of descriptors ($n$) across modalities, 
and the need to synchronize sequential information, we used a Long Short Term Memory (LSTM) network, followed by an \textit{AdaptiveAvgPool1d} layer and a \textit{Linear} fully connected (FC) layer to standardize temporal dimension $n$ to $n=50$ for all modalities and video clips.
During individual modality processing, the final layer was used to project the textual feature vector from $D_{T}=768$ to $D_{T}=384$, aiming to mitigate potential information redundancy. For the video modality, the model was fed the concatenation of flow and RGB features; after projection through the final linear layer, the resulting vectors had a dimension corresponded to $D_{V}=1024$.

\subsection{The MexHat Dataset}
Pretrained partitions for train, test and validation were created aimed to provide a reference for comparable and reproducible research\textcolor{red}{\footnote{MexHat dataset is available at: https://github.com/iltocl/MexHat.git}}.
Table~\ref{table: dataset-partitions} provides a summary of the data splits. 

\begin{table}[h!]
\caption{Final dataset partitions. For the three-class labeling column \textbf{NN}: no negative content, \textbf{OF}: offensive content and \textbf{HS}: hate-speech content. For the fine-grained labeling, if a video clip belongs to the HS class the subgroup is delimited by \textbf{hs1}: racism, xenophobic factors, \textbf{hs2}: sexism and LGBTQ+ related factors; and \textbf{hs3}: others.} \label{table: dataset-partitions}
\centering
\begin{tabular}{|l|c|r|r|r|r|r|r|}
\hline
\textbf{Partition} & \textbf{Number of video clips} & \textbf{NN} & \textbf{OF} & \textbf{HS} & \textbf{hs1} & \textbf{hs2} & \textbf{hs3} \\
\hline
test  & 278 & 116 & 123 & 39 & 6 & 15 & 18 \\
train & 824 & 394 & 314 & 116 & 28 & 47 & 41 \\
val   & 276 & 130 & 99 & 47 & 10 & 9 & 28 \\
\hline
\textbf{ALL}   & \textbf{1378} & 640 & 536 & 202 & 44 & 71 & 87 \\
\hline
\end{tabular}
%\label{table: dataset-partitions}
\end{table}

\begin{figure}[h!]
\centering
% --- IMAGE A ---
\begin{subfigure}[b]{0.45\textwidth}
    \centering
    \includegraphics[width=\textwidth]{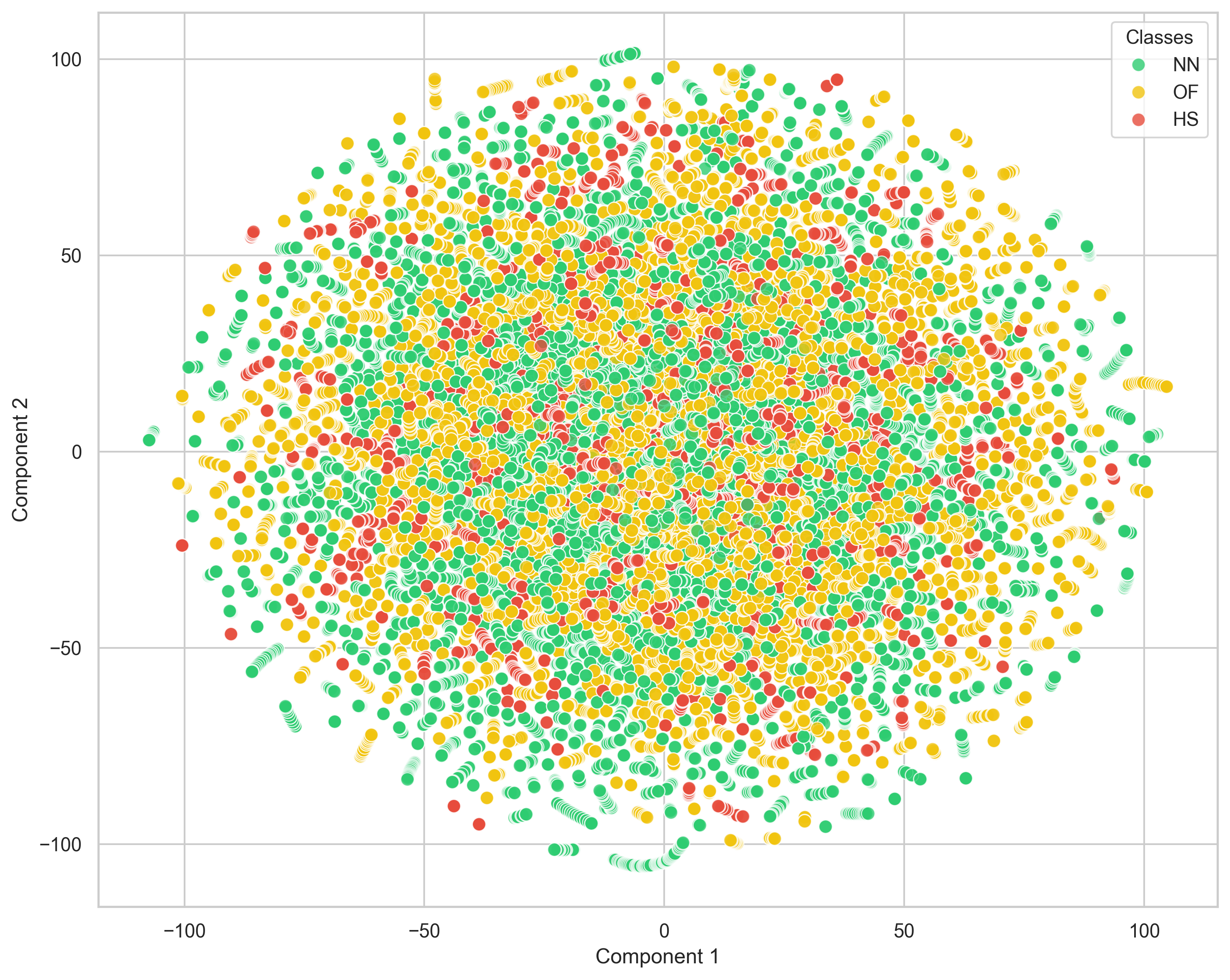}
    \caption{$T$}
    \label{fig:visualization-t}
\end{subfigure}
\hfill 
% --- IMAGE B ---
\begin{subfigure}[b]{0.45\textwidth}
    \centering
    \includegraphics[width=\textwidth]{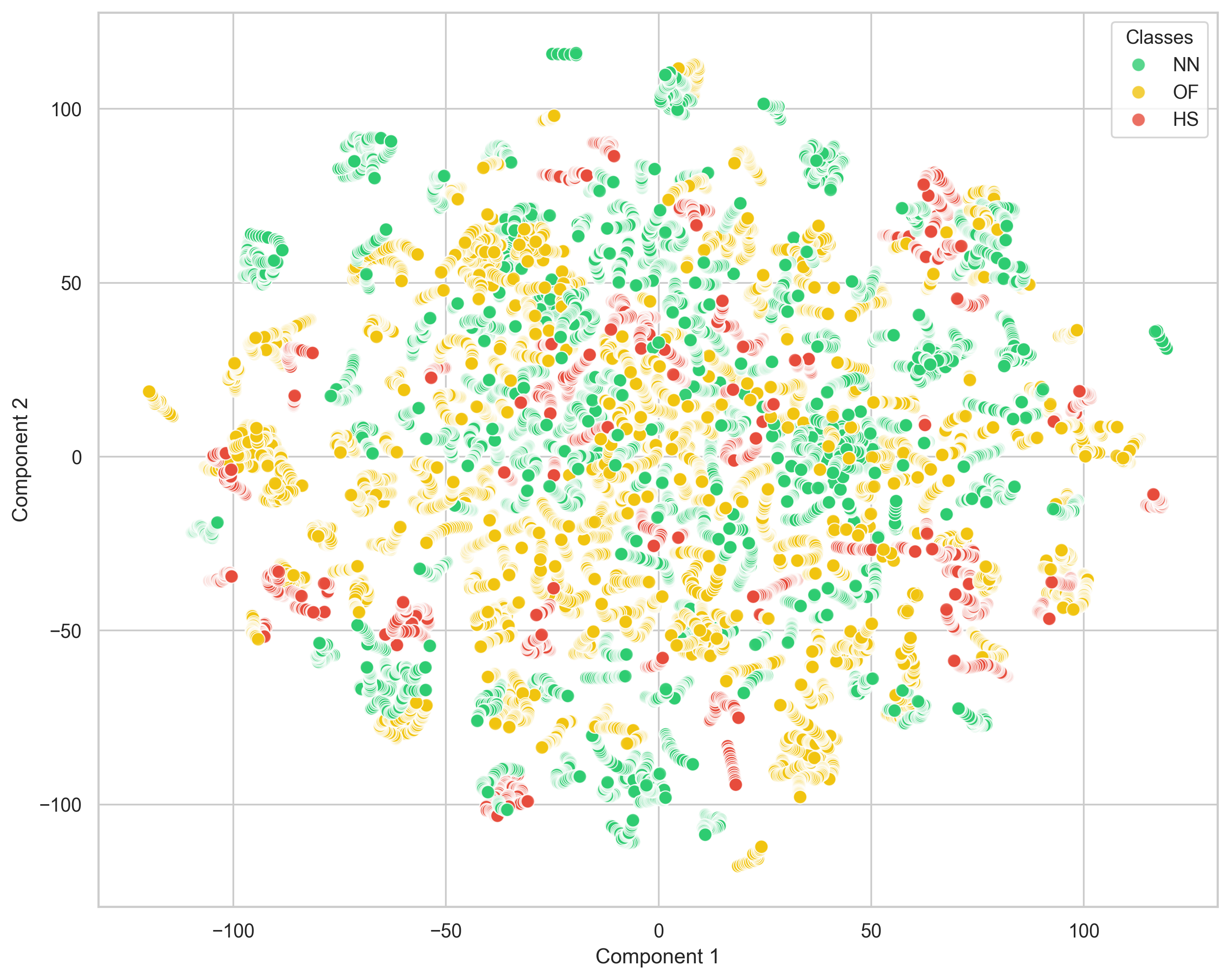} 
    \caption{$V$}
    \label{fig:visualization-v}
\end{subfigure}
\\
% --- IMAGE C ---
\begin{subfigure}[b]{0.45\textwidth}
    \centering
    \includegraphics[width=\textwidth]{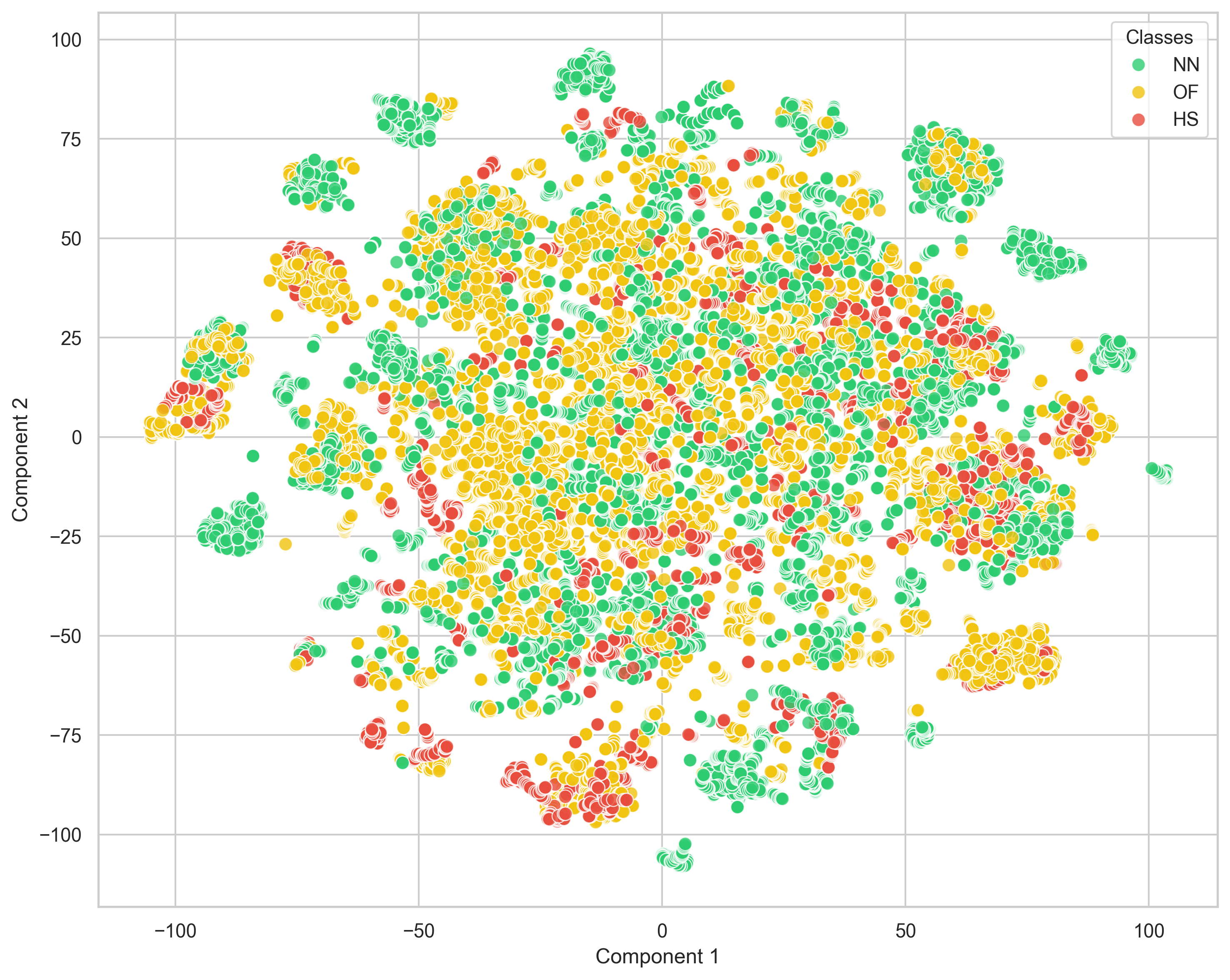}
    \caption{$A$}
    \label{fig:visualization-a}
\end{subfigure}
\hfill 
% --- IMAGE D ---
\begin{subfigure}[b]{0.45\textwidth}
    \centering
    \includegraphics[width=\textwidth]{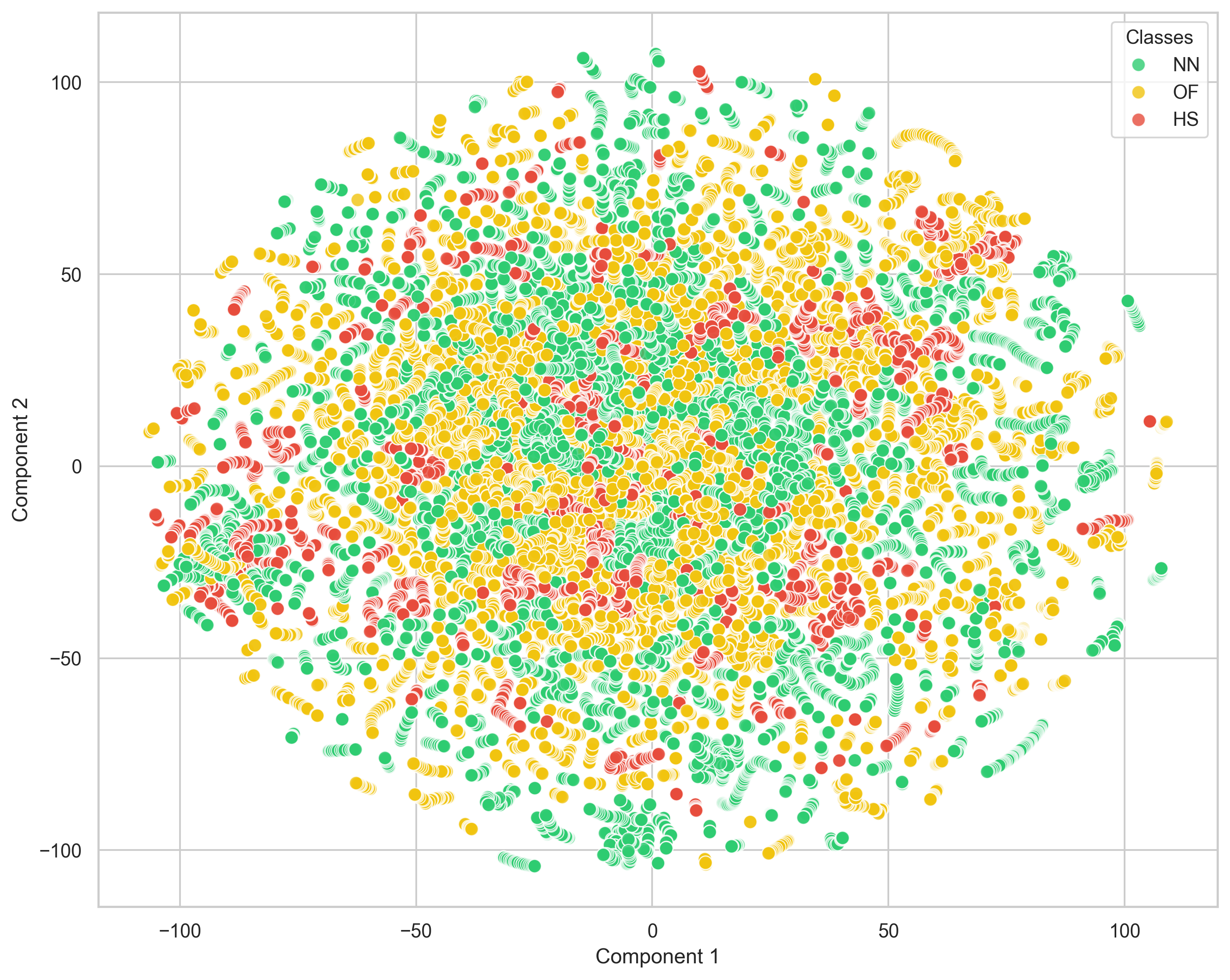} 
    \caption{$T + V+A$}
    \label{fig:visualization-tva}
\end{subfigure}
\caption{T-SNE visualization of data points for the test partition. Each point is a vector from each video representation and the color represents its associated class where NN: green, OF: yellow and HS: red; and $+$ refers to concatenation.}
\label{fig: dataset-visualization-tsne} 
\end{figure}

Using data from test partition, Figure \ref{fig: dataset-visualization-tsne} shows the data distribution between instances of three different classes into the same shared space. That is, instances were mapped into a two-dimensional space through applying T-SNE \cite{JMLR:v9:vandermaaten08a-tsne}.
Both unimodal (Figures \ref{fig:visualization-t}, \ref{fig:visualization-v} and \ref{fig:visualization-a}) and multimodal distributions (Figure \ref{fig:visualization-tva}) show a slight separation between classes NN, OF and HS. Reflecting the complexity of the associated learning problem.

\section{Experiments \& Results}
This section describes the supervised learning task associated to the MexHat benchmark and report experimental results. 
\subsection{Tasks}
Given a video $V$, we formulate two tasks. 
Task 1 corresponds to a three-way classification problem, where each video should be classified into \textit{no-negative}, 
offensive 
or hate-speech. 
Whereas task 2 (fine-grained classification) aims at explicitly recognizing the hate-speech variation; that is, groups $1, 2$ and $3$ (see section~\ref{subsection: annotation process}).

\subsection{Baselines}
We report baseline experimental results on the proposed MexHat dataset. 
Our evaluations include: 1)
\textbf{Uni-modal} representations from \textbf{T}ext, \textbf{V}ideo or \textbf{A}udio;
2) \textbf{Bi-modal} representations ($M_{1} + M_{2}$). In this case $+$ refers to concatenation;
and 3) \textbf{Multi-modal} representations of all modalities ($M_{1} + M_{2}+ M_{3}$).
The representations are fed into the classification model. 
We report accuracy and macro-f1 score as the main evaluation metrics for both tasks.

\subsection{Learning Models}
\textbf{Multi Layer Perceptron (MLP)}: An MLPClassifier from scikit-learn with \textit{relu} as activation function and \textit{adam} solver. 
For each video, all its associated $m$ temporally distributed feature vectors are independently classified.
Predictions for each class are counted and the final label corresponds to the majority one.
In case of drawn, one is randomly selected assigning the same probability.

\textbf{Transformer (TR) + MLP}: A Transformer Encoder from the Multizoo toolkit\footnote{\url{https://github.com/pliang279/MultiBench}}. The parameters of the model were as follows:  $nhead=\left [4 \right ]$, \textit{adam optimizer}, $epochs=100$, $learning\_rate=1e-5$,  $weight\_decay=0.01$, \textit{CrossEntropyLoss}, 
$nlayers=6$, and an MLP with two \textit{fully connected layers}, one \textit{dropout layer}, a \textit{ReLU} activation function and $output=\left [3,5 \right ]$ were fed with a concatenation $+$ of vectors from the three available modalities (\textbf{T}ext, \textbf{V}isual, and \textbf{A}udio).

\subsection{Results}
\label{subsection: results}
Three-way results shown in Table~\ref{tab: results three-class votes} 
show accuracies below $0.5$ excepting for rows 1 ($T$) and 6 ($A + T$) configurations. 
For macro-f1 scores only row 6 ($A + T$) achieves a slightly higher score than $0.4$. 
This evidences the complexity to identify patterns that allow a good discrimination between classes.
Particularly, columns F1-NN, F1-OF and F1-HS confirm that there is easier to identify \textit{no-negative} content followed by the \textit{offensive} class.
Considering the class imbalance few configurations were able to identify instances of the \textit{hate-speech} class. 

\begin{table}[h!]
\caption{Result from evaluating the unimodal, bimodal and multimodal representiations for the three-class classification.}
\label{tab: results three-class votes}
\centering
\begin{tabular}{|c|c|c|c|c|c|c|}
\hline
\textbf{Modality} & \textbf{Model} & \textbf{Acc} & \textbf{F1-macro} & \textbf{F1-NN} & \textbf{F1-OF} & \textbf{F1-HS} \\
\hline
T & MLP & \textbf{0.543} & 0.391 & 0.599 & 0.574 & 0\\
V & MLP & 0.468 & 0.363 & 0.533 & 0.475 & 0.080\\
A & MLP & 0.450 & 0.329 & 0.470 & 0.516 & 0\\
\hline
T + A & MLP & 0.489 & 0.353 & 0.531 & 0.528 & 0 \\
V + T & MLP & 0.435 & 0.335 & 0.523 & 0.407 & 0.075 \\
A + T & MLP & 0.568 & \textbf{0.423} & \textbf{0.603} & \textbf{0.616} & 0.049\\
\hline
A + V + T & MLP & 0.424 & 0.359 & 0.436 & 0.470 & \textbf{0.172}\\
\hline
T + V + A & TR+MLP & 0.460 & 0.327 & 0.532 & 0.450 & 0\\
\hline
\end{tabular}
\end{table}

\begin{table}[h!]
\caption{Result from evaluating the unimodal, bimodal and multimodal representiations for the fine-grained class classification.}
\label{tab: results fine-grained class votes}
\centering
\begin{tabular}{|c|c|c|c|c|c|c|c|c|}
\hline
\textbf{Modality} & \textbf{Model} & \textbf{Acc} & \textbf{F1-macro} & \textbf{F1-NN} & \textbf{F1-OF} & \textbf{F1-hs1}& \textbf{F1-hs2}& \textbf{F1-hs3} \\
\hline
T & MLP & 0.518 & 0.223 & 0.586 & 0.527 & 0 & 0 & 0\\
V & MLP & 0.442 & 0.227 & 0.525 & 0.419 & 0 & 0.105 & 0.087\\
A & MLP & 0.475 & 0.225 & 0.461 & 0.554 & 0 & 0.111 & 0\\
\hline
T + A & MLP & \textbf{0.522} & \textbf{0.247} & 0.568 & 0.549 & 0 & \textbf{0.118} & 0\\
V + A & MLP & 0.482 & 0.242 & 0.527 & 0.516 & 0 & 0 & \textbf{0.167}\\
A + T & MLP & 0.496 & 0.214 & 0.536 & 0.531 & 0 & 0 & 0\\
\hline
A + T + V & MLP & 0.475 & 0.204 & 0.548 & 0.470 & 0 & 0 & 0\\
\hline
T + V + A & TR+MLP & 0.417 & 0.118 & \textbf{0.589} & 0 & 0 & 0 & 0\\
\hline
\end{tabular}
\end{table}

Fine-grained classification results are reported in Table~\ref{tab: results fine-grained class votes}. Overall, the performance drops drastically. 
As task 1, most of the accuracy scores are below $0.5$ excepting row 1 ($T$) and row 4 ($T+A$) configurations.
Being more classes in the scenario F1-macro scores downgraded around $0.2$ being row 4 the best achieved score with $0.247$.
Per-class evaluations also show the prevalence of better performance for \textit{no-negative} content.
Unexpectedly, non of the configurations was able to identify sub category 1 of hate-speech (hs1). This group belonged to racism and xenophobic related factors.
For group 2 (hs2), three configurations (rows 2, 3, 4) showed some attempts to identify instances related with sexism and LGBTQ+ related factors.
Also, for group 3 (hs3) involving visual information in configurations $V$ and $V+A$ seem to contribute to identify types of hate-speech that not belong to the previous groups.

To sum up, in both evaluated tasks \textbf{T}ext only and the combination of \textbf{T}ext and \textbf{A}udio seem to be the better representations to capture cues that distinguish between classes.
Nevertheless, it is clear that by the low achieved scores, the general \textit{hate-speech} detection task requires strategies adapted to mitigate the class imbalance related to the real world scenario.

\subsection{Classification Errors}
\begin{figure}[t]
    \centering
    \includegraphics[width=0.9\linewidth]{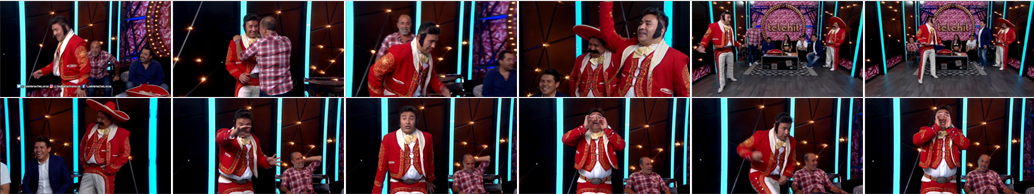}\\
    \tiny \textbf{Transcription:} Wait up, wait up.	I love you forever, dude.	How did the Mudo answer? Let's see, do it again, please.	Mudo! How? Good, good. There goes the Mudo.	He changed places.	He changed places! What's your name, Mudo? Look, you're going to stay here and keep watch.	When we hit land, you yell at all of us, because there are some girls out there and we all want to hit that.	We're going to sleep for a bit.	And after an hour, the Mudo goes...	Ti, ti, ti, ti, ti! Ti, ti, ti, ti! Screw it! Everyone jumps in.	"Oh motherfucker, damn Mudo! You stay right there. And then everyone starts talking at once."	Holy shit, the Mudo starts doing it! Ti, ti, ti, sharks! Run for your lives!
    \caption{Video instance (0tZ5s2W6nZY.05) wrongly classified.}
    \label{fig:error_classification_video}
\end{figure}

\begin{figure}[h!]
    \centering
    \includegraphics[width=0.8\linewidth]{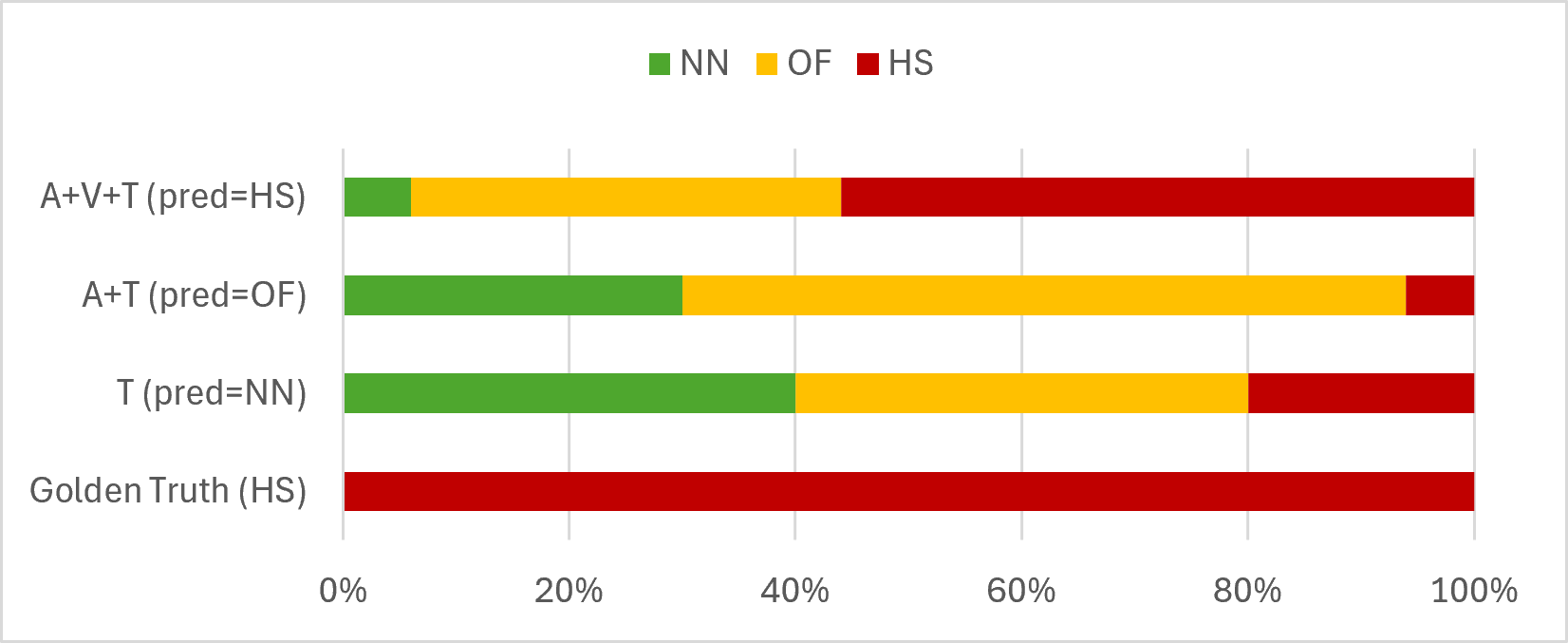}
    \caption{Video instance (0tZ5s2W6nZY.05) wrongly classified.}
    \label{fig:3classes_error_classification_video}
\end{figure}

Evaluating at vector level allows us to visualize the percentage of class decisions for a determined video instance.
The video instance shown in Figure \ref{fig:error_classification_video} shows a situation where textual, audio (tone of the joke) and visual (corporal language) modalities could be considered required to understand the whole intention.
The golden truth of this 
corresponds to \textit{hate-speech} content (Figure \ref{fig:3classes_error_classification_video}). When evaluating it with the unimodal ($T$, MLP with votes count) configuration $40\%$ of its content was classified as \textit{no negative}, $40\%$ as \textit{offensive} and only $20\%$ as HS. Because in the case of draw a random choice is done, it ended up predicted as NN.
For the case of bimodal ($A+T$, MLP with votes count) \textit{best} configuration for macro-F1 (See Table \ref{tab: results three-class votes}) $50\%$ of its content was labeled as \textit{offensive}, a class closer to \textit{hate-speech} but still the HS veredict was minimum.
Finally, the use of the three modalities ($A+V+T$, MLP with votes count) indeed identified most of the content into the HS class followed by the OF class and a minimum percentage of \textit{no negative} content. This configuration got the prediction right.
This way of taking advantage of the video content analysis could provide us with directions on how to mitigate misclassification at video level by searching for ways to focus on certain cues at a more detailed level. 

\section{Conclusions}
We introduced a new multimodal video dataset aimed for the hate-speech detection task for Mexican Spanish content.
We presented the followed pipeline to the construction of our dataset by detailing the stages for video data collection, data annotation, feature extraction, and baseline evaluations.
The evaluation stage involved two ways to perform the hate-speech detection task. First, as a three class classification problem considering classes: \textit{no negative content}, \textit{offensive content} and \textit{hate-speech content}. Second, as a fine-grained class classification problem where three sub groups of hate-speech were defined.
To establish a performance baseline, we evaluated a variety of unimodal, bimodal, and multimodal representations across both tasks.
The experimental results confirmed how hate speech is bound to culture and context in order to be interpreted. Particularly, in the Mexican scenario where the linguistic cues and expressions create a very thin line between harmless humor and offensive intentions.
Our findings are summarized as follows:
\begin{itemize}
    \item Baseline results showed that text, either alone or combined with other modalities, provides sufficient information to achieve strong initial performance.
    \item While combining visual and audio cues achieves good performance, accurately distinguishing hate speech remains a significant challenge.
    \item Misclassified instances suggest that in order to understand the speaker's intention an adequate integration of multiple cues (e.g., linguistic with text and body language with visual modality) is required.
    \item Vector-level evaluation provides a feasible method for tracking down classification errors, suggesting that a more precise capture of specific evidence could benefit the global decision.
\end{itemize}
Given the inherent class imbalance, as future work, adjusting evaluations for adapted penalizations would be required to boost performance.

\begin{credits}

\subsubsection{\ackname}Itzel Tlelo was supported by SECIHTI (scholarship: 972915).

%\subsubsection{\discintname}
%It is now necessary to declare any competing interests or to specifically state that the authors have no competing interests. Please place the statement with a bold run-in heading in small font size beneath the (optional) acknowledgments\footnote{If EquinOCS, our proceedings submission system, is used, then the disclaimer can be provided directly in the system.}, for example: The authors have no competing interests to declare that are relevant to the content of this article. Or: Author A has received research grants from Company W. Author B has received a speaker honorarium from Company X and owns stock in Company Y. Author C is a member of committee Z.
\end{credits}
\bibliographystyle{splncs04}
\bibliography{references.bib}

\end{document}